*Pre-print*

# Mapping Historic Urban Footprints in France: Balancing Quality, Scalability and AI Techniques

**Walid Rabehi [1*], Marion Le Texier [2] and Rémi Lemoy [3,]**

[1] CY Cergy Paris Université/ PLACES Lab, France; walid.rabehi@cyu.fr
[2] University of Montpellier 3/ LAGAM Lab, France ; marion.le-texier@univ-montp3.fr
[3] University of Rouen/UMR IDEES Lab, France ; remi.lemoy@univ-rouen.fr

[*] Correspondence: walid.rabehi@cyu.fr;

**Abstract**

Quantitative analysis of historical urban sprawl in France before the 1970s is hindered by the lack of nationwide digital urban footprint data. This study bridges this gap by developing a scalable deep learning pipeline to extract urban areas from the Scan Histo historical map series (1925-1950), which produces the first open-access, national-scale urban footprint dataset for this pivotal period. Our key innovation is a dual-pass U-Net approach designed to handle the high radiometric and stylistic complexity of historical maps. The first pass, trained on an initial dataset, generates a preliminary map that identifies areas of confusion, such as text and roads, to guide targeted data augmentation. The second pass uses a refined dataset and the binarized output of the first model to minimize radiometric noise, which significantly reduces false positives. Deployed on a high-performance computing cluster, our method processes 941 high-resolution tiles covering the entirety of metropolitan France. The final mosaic achieves an overall accuracy of 73%, effectively capturing diverse urban patterns while overcoming common artifacts like labels and contour lines. We openly release the code, training datasets, and the resulting nationwide urban raster to support future research in long-term urbanization dynamics.

## 1. Introduction

Capturing long-term spatio-temporal dynamics of urban sprawl is critical for modeling anthropogenic dynamics such as urban growth and soil imperviousness [1].

In France, nationwide cartographic coverages begin with the Cassini maps in the 18th century, the Etat-Major maps in the 19th century, and the Scan Histo series in the 20th century [2]. However, a lack of quantitative analyses at the national scale (e.g., concerning surface area, orientation) persists for these periods, especially when compared to the current era, which starts in the 1970s with the advent of the Landsat archives [3], followed by the SPOT archive in 1986 [4]. This gap is mainly due to the nature of the data, which consists of scanned maps. For GIS applications, such maps require classification and conversion into more readily exploitable formats (e.g., shapefile, GeoPackage) [5]. This conversion process remains the most problematic step, owing to the inherent complexity of historical maps [6].

Thus, generating a digital coverage of urban areas in France for the 20th century, using the Scan Histo 1950 collection provided by the French National Institute of Geographic and Forest Information (IGN) [7], is a critical need. This historical period is marked by World War II, involving destruction and subsequent reconstruction [8], as well as by the expansion of automobile use in its early decades [9].

Comparing these temporal footprints with numerical data from the 1970s—when the satellite era began and systematically classified data became available—provides a highly original source for bridging this temporal gap.

The automated detection of urban features from historical cartographic documents represents a critical challenge in geospatial sciences, as it is particularly hindered by the intrinsic heterogeneity of the urban fabric and the complexity of spatial arrangements. Early approaches, which rely on manual digitization or threshold-based techniques [10][11], often prove inadequate when confronted with the stylistic diversity and paper degradation of pre-20th-century maps, such as France's Cassini or Napoleonic cadasters [12][13]. These limitations spur growing interest in deep learning–based methods, which demonstrate greater robustness to noise and variability [14].

Studying historical urbanization is essential for modeling long-term anthropogenic impacts [15], yet consistent urban footprint data remain scarce before the satellite era. France's Scan Histo archive (1925–1950) provides a unique opportunity to address this gap, but no prior work leverages its nationwide coverage for automated urban extraction.

Recent breakthroughs in convolutional neural networks (CNNs), particularly U-Net architectures [16], enable high-accuracy feature extraction from historical maps. For instance, [17] achieves more than 85% precision in segmenting 19th-century road networks, while [18] demonstrates the feasibility of applying these methods to old French maps of the 18th century for natural classes. However, these studies focus either on localized case studies or on feature types other than urban footprints, leaving broader temporal and spatial scales unexplored.

Other approaches for object detection, such as Fully Convolutional Networks (FCN) [19] or Artificial Neural Networks (ANN) based on pixel-to-pixel detection [20], have also been explored. However, such methods are often less effective in urban areas marked by greater heterogeneity, which typically require very large datasets and sensitive parameter tuning. This limitation justifies the growing interest in convolutional neural network approaches.

Among these, popular semantic segmentation architectures such as DeepLabV3+ gain attention, particularly in satellite imagery [21]. Nevertheless, [22] highlights that this architecture still requires improvements—particularly through transfer learning and active learning—to enable large-scale reproducibility.

A recent contribution [23] demonstrates that the majority of deep learning approaches, even when oriented toward urban detection, primarily focus on active and passive satellite imagery, with only a few addressing large cartographic scales. Most Deep Learning (DL) contributions related to historical maps at national scales focus more on forest and agricultural land use [24][25]. In France, the only large-scale contribution to date [26] reveals persistent challenges, such as confusion between roads and textual elements, and a lack of openly accessible results, including code, datasets, and data.

The lack of large-scale DL approaches for urban footprints on historical maps can be attributed to several factors. These include the initial scarcity of map coverage before the 18th century, a period where maps primarily depicted major capital cities [27]. Moreover, the inherent complexity and variable preservation quality of these maps introduce challenges that are often more complex than those in satellite imagery, such as mosaicking with heterogeneous color tones due to temporal variability and different producers, ink degradation, and structural variability of patterns [28][29].

This paper presents a scalable deep learning framework to extract urban footprints from the Scan Histo collection (early 20th century), addressing three key innovations:

- A double U-Net variant optimized for heterogeneous map styles, capable of handling elements like hatching, text labels, and marginalia, where the first application uses native maps and the second uses a medium binary mapping.

- An open-access nationwide urban database for France's mid-20th-century transition period.

- A full open-access release of training data, code, and vectorized outputs to support reproducibility.

By bridging historical cartography and modern AI, our work enables the systematic analysis of urbanization patterns prior to the 1970s—a period preceding widespread remote sensing data that remains poorly documented in diachronic and spatial analysis due to the lack of digital urban footprints.

## 2. Materials and Methods

This approach is developed following several inconclusive tests using traditional remote sensing methods, including pixel-based classifiers, hierarchical classifications, and object-based image analysis [30]. Additionally, the straightforward application of a standard CNN architecture, such as U-Net, DeepLab, or SegNet, also demonstrates significant limitations at large scales [26].

The method involves applying a U-Net architecture (Figure 1) with a highly specific approach:

A first general prediction is performed using a small dataset of 58 image/prediction pairs. This step produces an initial raster map of France that reveals global errors and identifies areas of high complexity.

- A second dataset is then created (with targeted sampling of previously detected problematic zones), increasing the dataset to 312 pairs (roads, text, contour lines (see next section tables). This second iteration generates a cleaner binary raster using 80% of the dataset have been used.

- In the third step, the intermediate raster from the previous stage is used (instead of the original scanned maps, using the same previous dataset pairs (Figure 1).

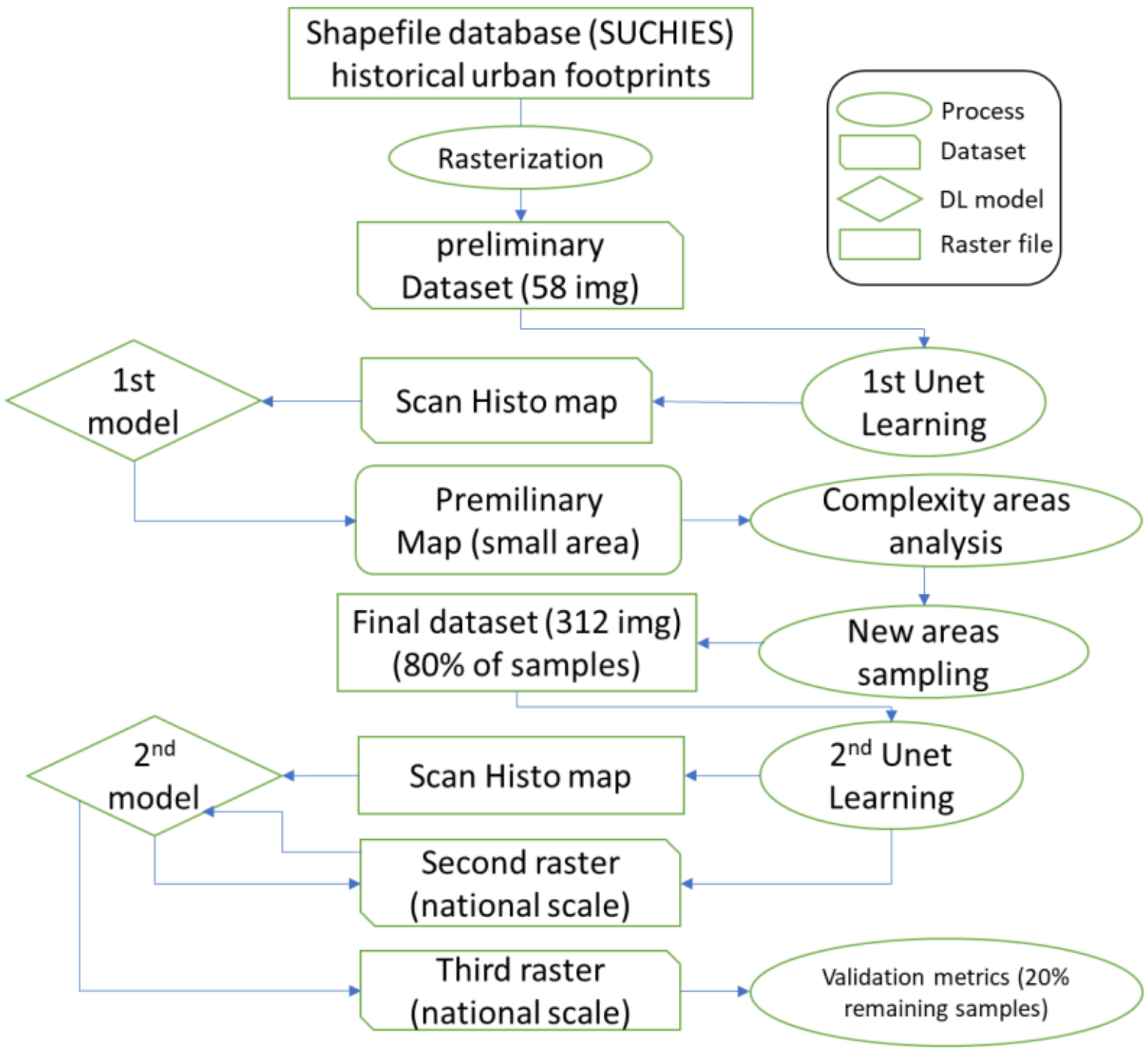

**Figure 1**. Overall methodology

In the last step and using the remaining 20% of datasets-pairs, accuracy metrics have been computed (and described in next sections).

In this contribution, we also argue that generative artificial intelligence (GenAI) has been used in this paper, specifically OpenAI's ChatGPT 3.5 (open access) and DeepSeek (open access), to assist in debugging and organizing the code.

### *2.1. Data Preparation*

The Scan Histo collection comprises digitized topographic maps produced by the French national institute "IGNF" between 1925 and 1950s [7]. These maps are georeferenced and available at high resolution, providing detailed representations of urban features, infrastructure, and landscape elements. The archive covers the entirety of France (excluding the overseas departments and Corsica) and is well-suited for spatial analysis and semantic segmentation (Table 1).

**Table 1.** data description

| | description |
|---|---|
| Data link | https://geoservices.ign.fr/scanhisto#telechargementscan501950 |
| Raster type | JP2, RGB, float92, 5 m resolution |
| | 5000x5000 pixels, 25x25 km |
| Raster size | 45 Mbytes (by tile), 941 Scene for France scale (53 G bytes) |

2.1.1. Urban types

The urban unit is differentiated on the maps through a center–periphery partitioning: downtown blocks are represented as rectangles with internal textures, whereas peripheral urban units are depicted as solid pixels. In some cases, differentiation is also achieved by region, with certain regions adopting black and others red (as in the case of Paris and its surrounding area), while still maintaining the previously mentioned center–periphery texture principle (Table 2).

**Table 2**. Urban footprint possible patterns

| Type | Overview | Description |
|---|---|---|
| urban unit type 1 | | City center blocks |
| urban unit type 2 | | Perpheral urban units |
| urban unit type 3 | | Countryside urban units |
| urban unit type 4 | | Urban units in Paris Area (Ile de France), block for city center, and full red polygones for perphery |

### 2.1.2. Data complexity

The complexity of implementing a predictive model stems not only from the previously discussed typology, which illustrates the various manifestations of the urban unit, but more significantly from other factors. These include temporal discrepancies between cartographic coverages, which were produced by different human operators, and physical deterioration caused by aging and storage conditions. Furthermore, the challenge of extracting urban features from historical scanned maps is compounded by the abundance of symbols rendered in black, such as contour lines, internal textures of vegetated areas, administrative boundaries, and text. This high density of features increases the potential for confusion during automated classification (Table 3).

**Table 3**. Visual variability

| Type | Overview |
| --- | --- |
| **Mosaique radiometric difference** | |
| **Black lines** | |
| **Qualitative symbols / Text similarity** | |

### 2.1.3 Datasets Description

The dataset consists of georeferenced RGB images, derived from historical maps, and corresponding binary masks that indicate urban areas (Figure 2). All images are resized to 256×256 pixels to ensure uniformity. The images undergo normalization, and the masks are binarized, where urban areas = 1 and non-urban areas = 0.

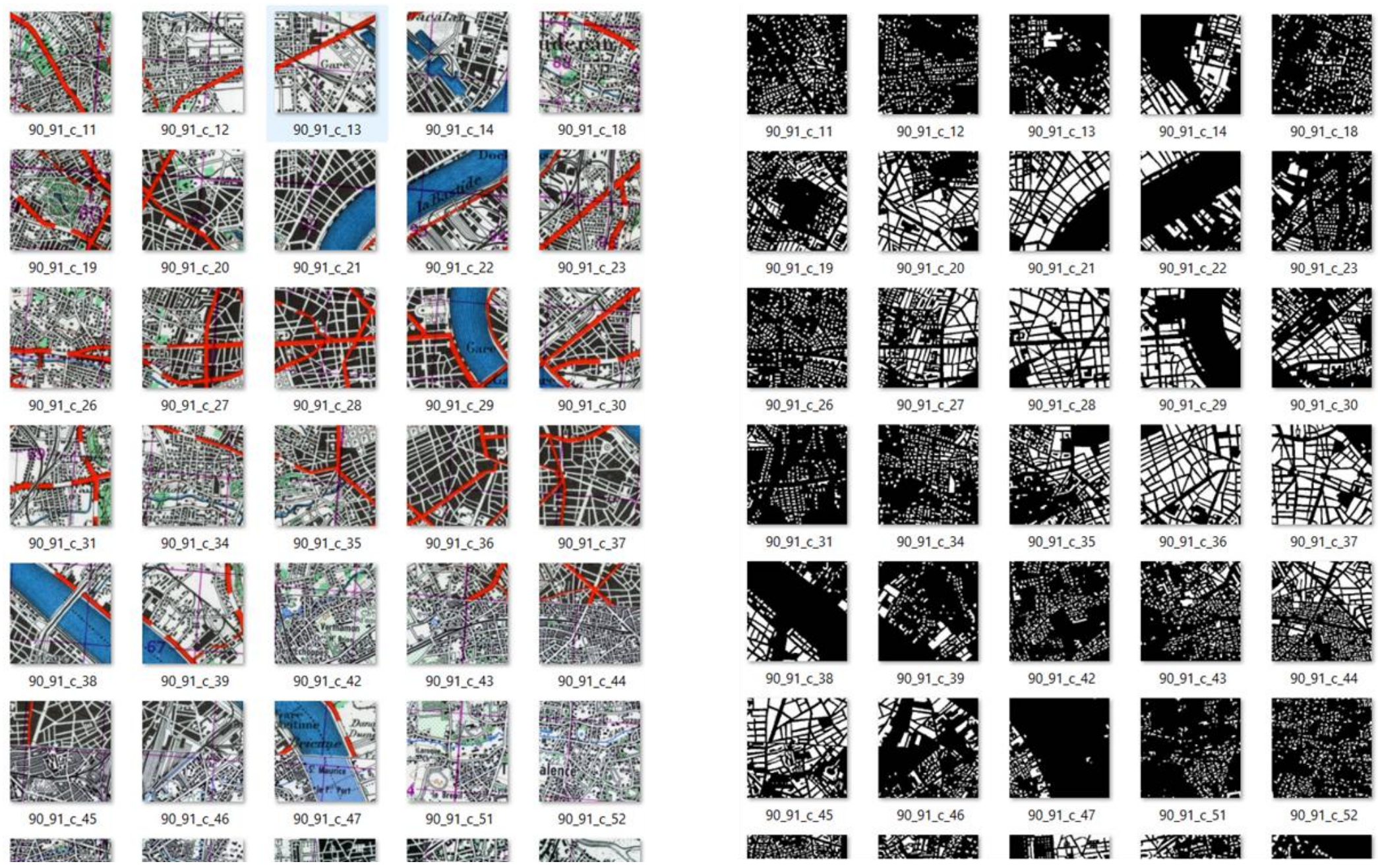

**Figure 2**. Openacess dataset (truth/reference footprints)

The native dataset is publicly available as of May 2024 and contains urban footprints for 84 French cities [31]. This shapefile database was created through manual digitization of 1–10 km buffer zones around cities, with the zone size defined by their relative importance. The datasets used in the present work are derived from this database. Both the datasets and all Python code are openly accessible via a Nakala repository (https://nakala.fr/10.34847/nkl.aea7388x) and a GitHub repository (https://github.com/wilius47/SUCHIES-deposit).

#### 2.1.4. Custom Dataset

A custom Dataset class is implemented using the PyTorch deep learning framework [32] to handle the loading, preprocessing, and transformation of both images and masks. This class ensures correct pairing and preprocessing for efficient model training.

### *2.2. Model Architecture*

#### 2.2.1. UNet Design and implementation

We adopt the U-Net convolutional neural network architecture due to its established effectiveness in semantic segmentation tasks [33][34]. The model is trained on the rasterized Scan Histo data to classify each pixel as urban or non-urban.

Our implementation uses a standard U-Net architecture comprising an encoder-decoder structure with skip connections [35]. The encoder progressively downsamples the input image, while the decoder upsamples the feature maps and reconstructs the spatial structure using information from the corresponding encoder layers.

#### 2.2.2. Implementation

The model is implemented in PyTorch with the following specifications:

- Input: RGB images (3 channels)
- Output: Binary mask (1 channel with sigmoid activation)
- Loss function: Binary Cross Entropy (BCE)
- Optimizer: Adam
- Training epochs: (from 1 to 30 epochs)

- Batch size: 8
- Data splitting: The labeled dataset was partitioned into training (80%) and validation (20%) subsets.

The processing pipeline (training, inference, and mosaicking) was executed on the Austral HPC cluster at CRIANN, using a MIG-enabled NVIDIA A100 GPU (12 GB partition) with 4 CPU cores and 10 GB RAM, under an 8-hour session limit (40 hours total)

*2.3. Training and Evaluation :*

The training process involves hyperparameter tuning, including adjustments to batch size and the number of iterations. The model's performance is evaluated using standard segmentation metrics such as pixel-wise accuracy, precision, recall, and the F1-score [36]. A second, more geometrically-oriented evaluation is conducted using the Intersection over Union (IoU) metric [37]. The optimal configuration is selected based on the highest quality segmentation results on the validation samples.
The model trains for 30 epochs on a workstation with optional GPU acceleration. During each epoch, the model optimizes its parameters to minimize the Binary Cross-Entropy (BCE) loss [38] between the predicted masks and the ground truth. Sample images and their corresponding predicted masks are visualized to monitor training progress. The segmentation metrics are formally defined in the equations below.

- Precision (P):

$$\frac{TP}{TP+FP} \tag{1}$$

- Recall (R):

$$\frac{TP}{TP+FN} \tag{2}$$

From these simplistic metrics, we can also compute:

- F1 Score (Dice coefficient): the harmonic mean of Precision and Recall

$$2\ x\ \frac{Precision\ x\ Recall}{Precision+Recall} \tag{3}$$

- Intersection over Union (IoU): known as the Jaccard Index

$$\frac{TP}{TP+FP+FN} \tag{4}$$

where TP= True Positives, FP = False Positives, and FN = False Negatives.

For optimal model parameters choice (epochs number):

Binary Cross Entropy Loss:

$$J_{bce} = \frac{1}{M}\sum_{m=1}^{M}[y_m \times \log\left(h_\theta(x_m)\right) + (1 - y_m) \times \log\left(1 - h_\theta(x_m)\right)] \tag{5}$$

Where M: number of training examples, ym: target label for training example m, xm input for training example m, and hθ: model with neural network weights θ [38].

*2.4. Application to Large-Scale Inference*

2.4.1. Tiling and Prediction

Once trained, the model is applied to the 941 Scan Histo map tiles covering the national territory. Each tile is subdivided into 256×256 pixel patches to optimize processing, and predictions are computed independently. A two-step mosaicking process reconstructs first the full tiles and then the complete urban footprint map for France.

Predicted mask tiles are merged into full mosaics, corresponding to the original tiles, using the JP2 input files as a spatial reference. All geospatial metadata, including the coordinate reference system (CRS) and geotransform, is preserved. Finally, all individual mosaics are merged to produce a comprehensive urban mask mosaic at the national level.

2.4.2. Second level training

To improve segmentation quality and reduce false positives from features such as roads, contour lines, or map labels, a second inference pass is applied. Instead of using the original RGB maps, this step takes the binary urban mask generated in the first pass as its input. This iterative refinement, which operates on a less heterogeneous input, enhances spatial coherence and effectively suppresses spurious detections.

## 3. Results

This study selects the U-Net architecture for its strong performance in semantic segmentation, particularly with limited training data. For comparison, we include DeepLabV3+, a popular deep learning model known for capturing multiscale contextual information [23], and K-means, a traditional unsupervised method that provides fast, label-free clustering [39]. We also evaluate a fourth, semi-automatic approach that combines radiometric thresholding with land-cover class filtering [40].

This comparative setup demonstrates the advantages of deep learning over classical methods in terms of both segmentation accuracy and spatial coherence (Figure 3).

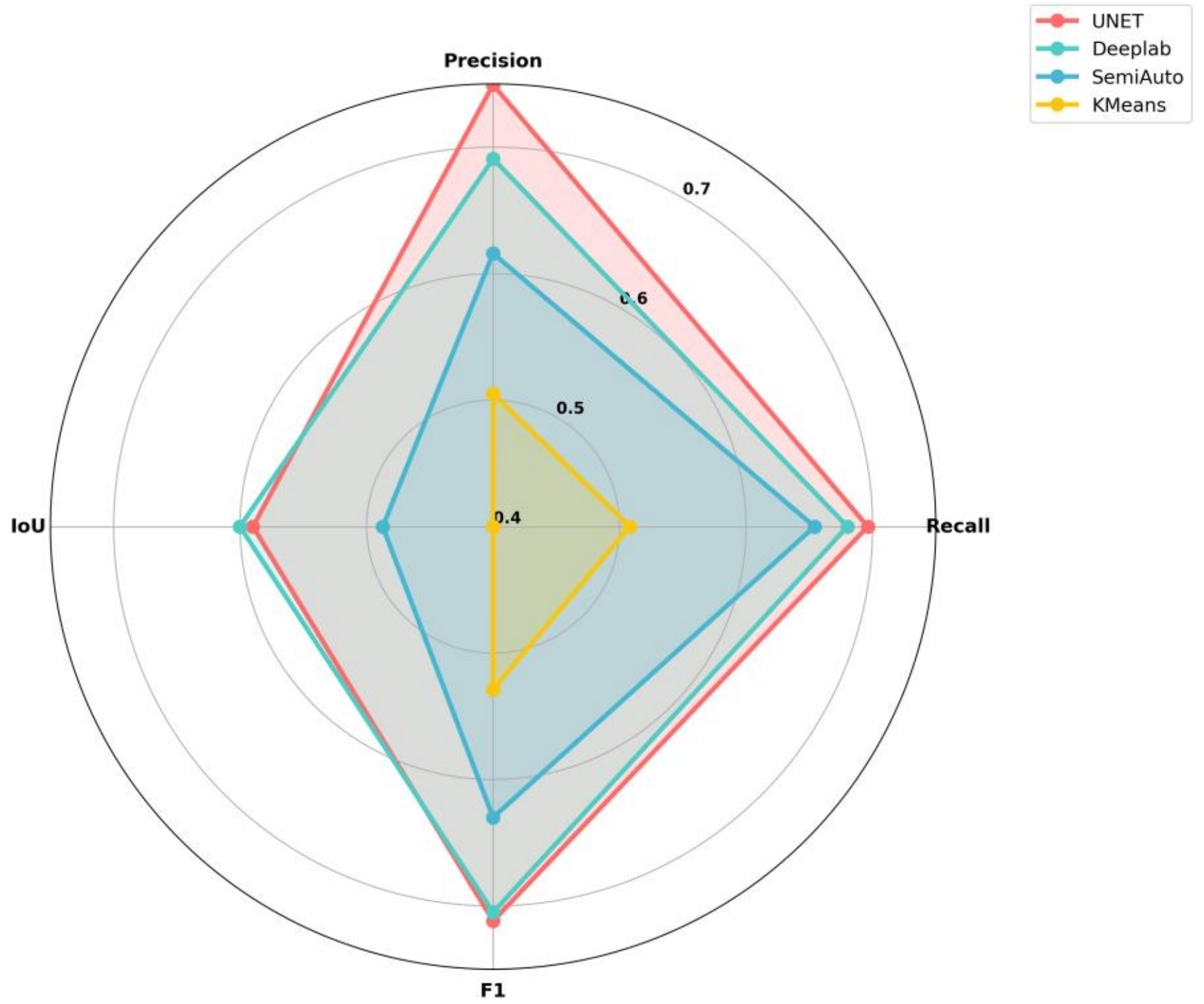

**Figure 3.** Comparaison with popular classification algorithm

The quality of the UNET raster (73% accuracy) clearly stands out from the two products derived from the “semi-automatic” method (empirical classification) and the "unsupervised classification (k-means)," which display the lowest rates (being purely radiometric). Nevertheless, Deeplab shows more acceptable rates (69%), still with a lesser quality (comparing to UNET), particularly in rural areas.

The first iteration on the native map detects successfully all fully black buildings (peripheral areas) and ignores small text from map labels (1, 3, 4 maps, Figure 4). However, it struggles with large-letter labels (e.g., city names in bold uppercase) and textured blocks (map 3). The algorithm also identifies hatched-outline urban blocks but fails to reconstruct their precise shapes with this initial dataset.

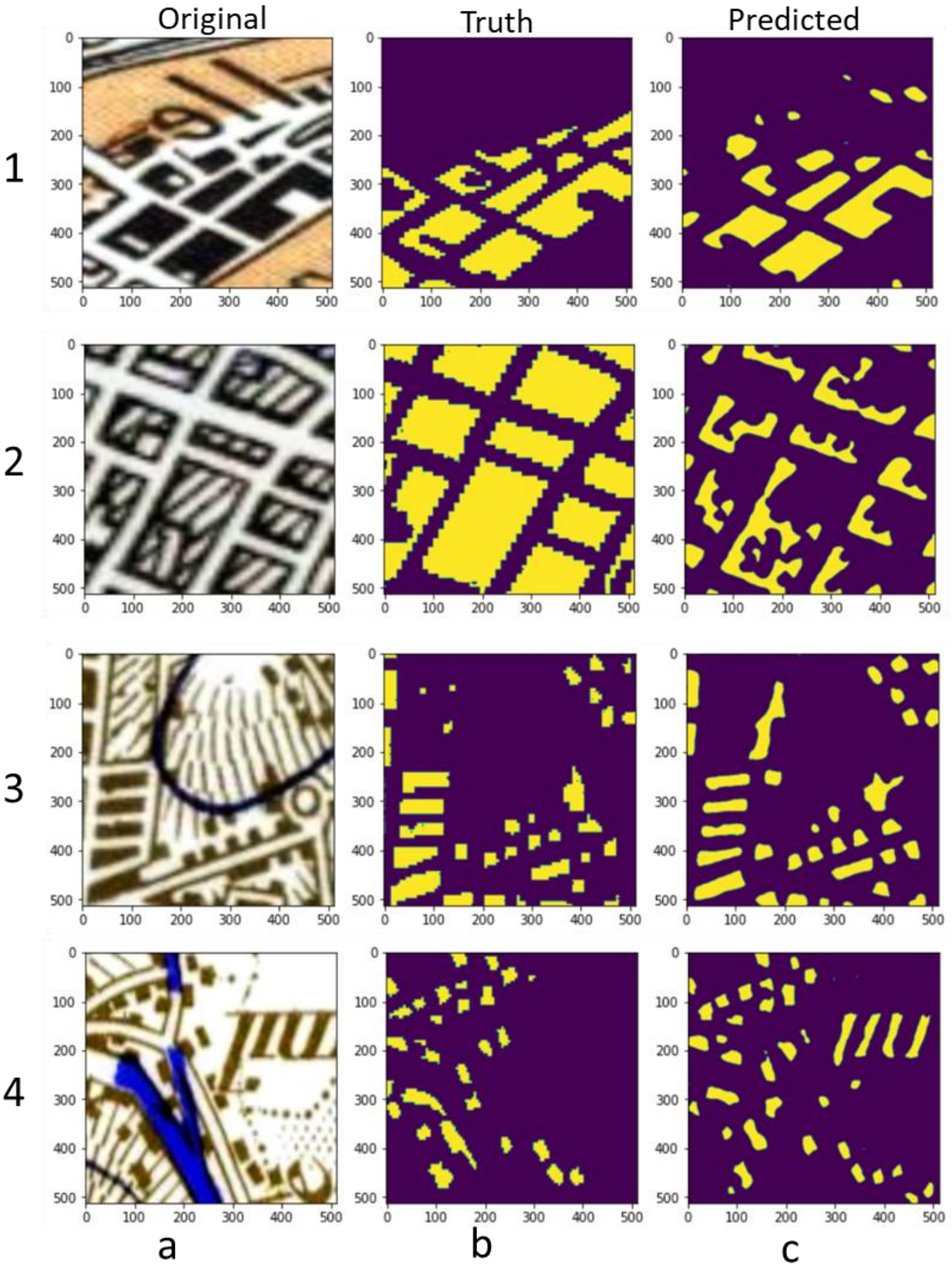

**Figure 4.** First prediction classification issues

The model successfully segments urban areas The end-to-end system is scalable and geospatially accurate. Some challenges remain, including misclassification in mixed land cover regions and variations due to maps conservation or scanning conditions.

Visual inspection confirms the model's effectiveness in capturing urban patterns across diverse regions. Quantitative performance metrics (IoU, Precision, Recall) are computed in the next steps of this contribution.

The second iteration (using a significantly larger dataset) enables the model to capture all urban forms and overcome inconsistencies in cartographic quality.

Subsequently, the focus shifts to optimizing the model's learning parameters. For this purpose, we employ an approach similar to that used for quality assessment, implementing an overall accuracy metric to prevent model overfitting (Figure 5).

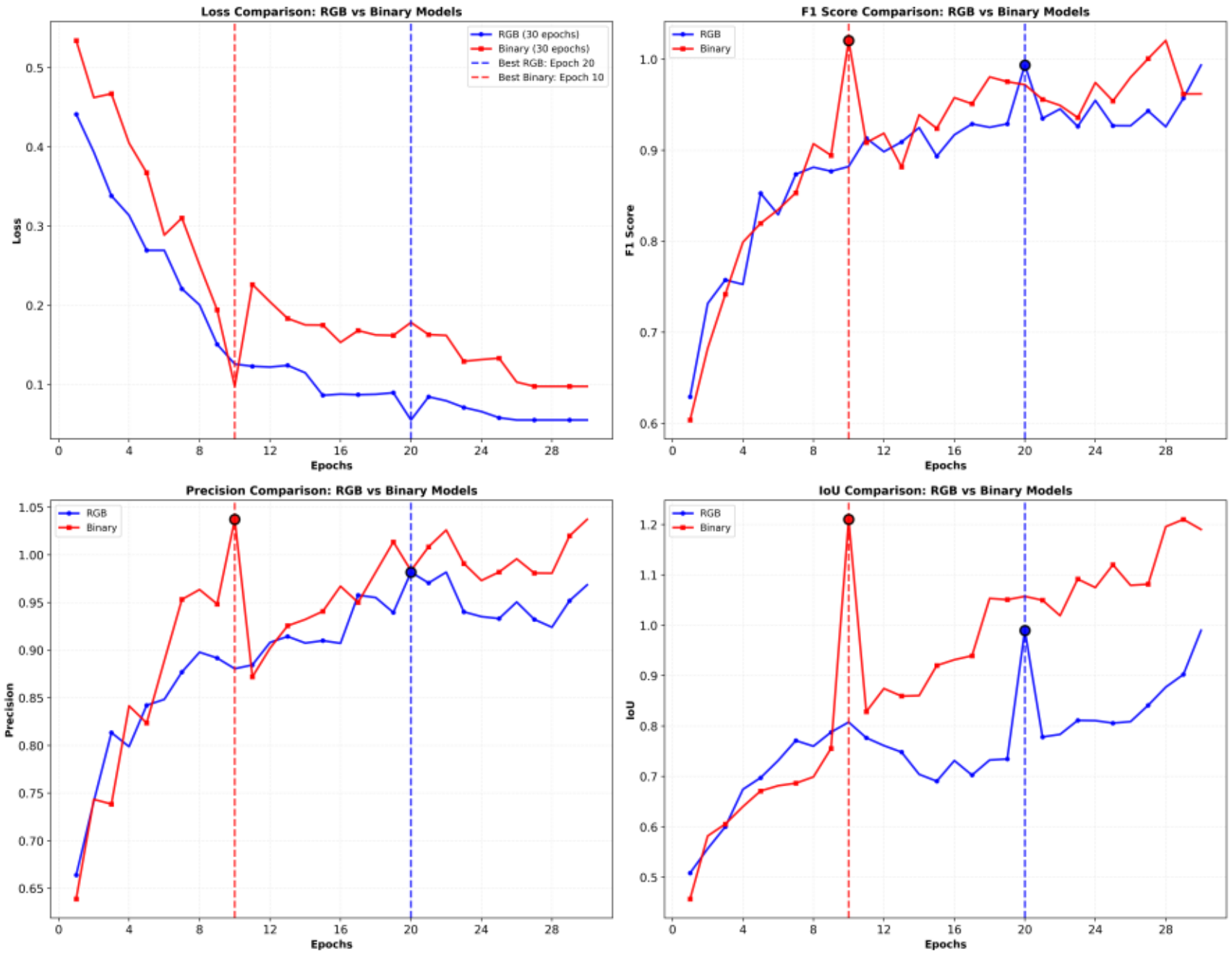

**Figure 5**. Analyzing optimal model Epochs "loss and f1 score for the two rasters"

The analysis reveals an optimal number of 20 epochs for the first raster, which is computationally expensive but logical given the high radiometric variability of the original image (colors, textures, hues, etc.), requiring greater processing effort. For the second iteration, 10 epochs were sufficient as the binary raster exhibited lower heterogeneity compared to the RGB version.

## 4. Discussion

The iteration analysis reveals that by the tenth iteration, the model successfully extracts urban areas in their entirety, though with several false positives such as roads and text, due to its high sensitivity. Between iterations 11 and 15, the model eliminates most text-based false positives (Figure 6). The final iteration phase further removes linear roads in addition to text, achieving the optimal sensitivity-precision trade-off. However, exceeding the 21st iteration leads to overfitting, which causes the model to overlook even clearly visible built-up areas.

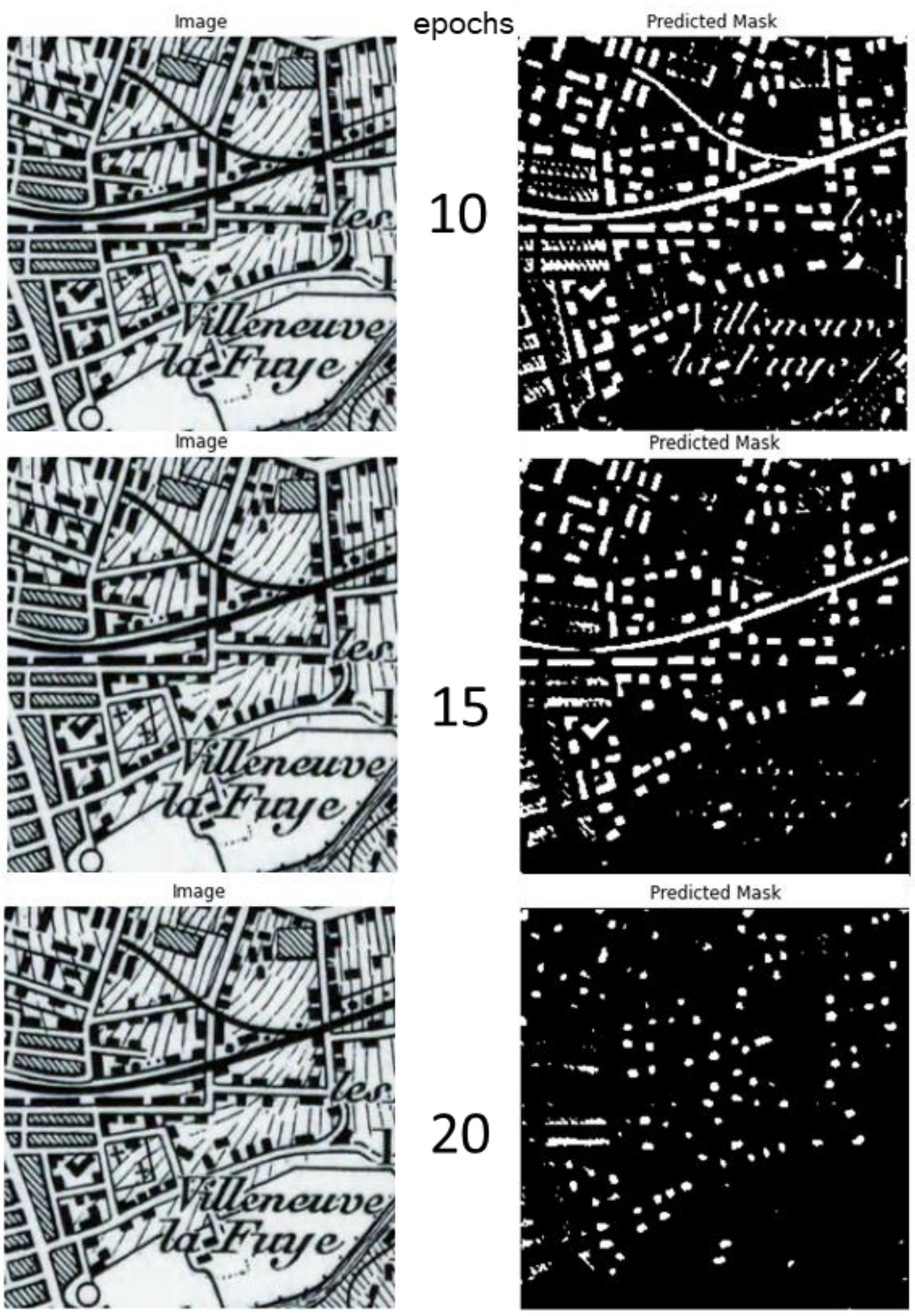

**Figure 6**. Overview by iteration in urban areas

In rural areas, especially those with steep terrain, false positives are very frequent (notably recurring contour lines, the texture of agricultural land, etc.), which still remain partially even after 20 additional iterations (Figure 7).

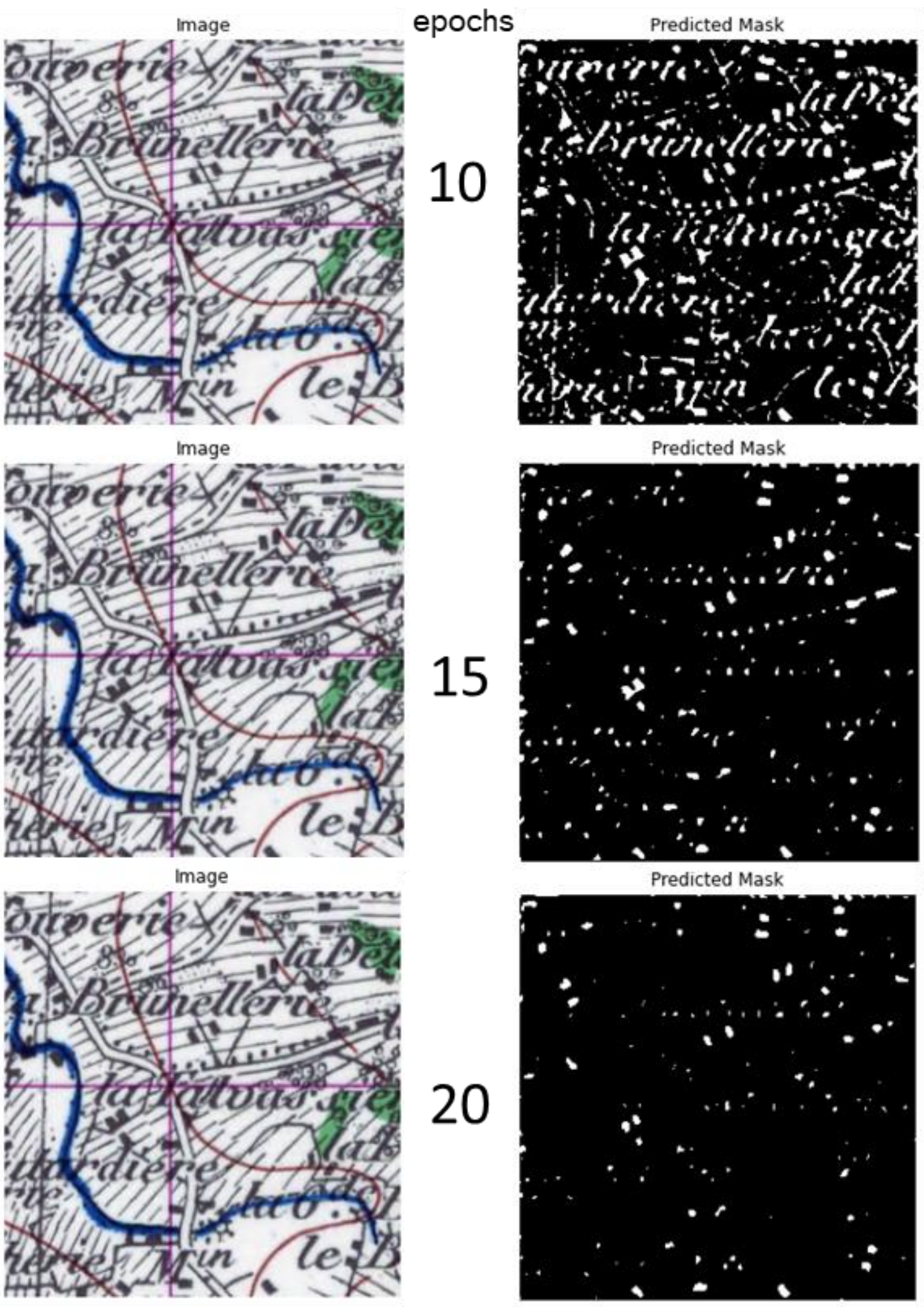

**Figure 7**. Overview by iteration in rural areas

The proposed method produces detailed urban footprint maps consistent with known settlement structures. Preliminary evaluations indicate a high level of agreement between predicted and manually digitized urban areas, with significant improvements after the refinement step. Remaining errors are primarily associated with heterogeneous map styles and ambiguous features.

The overall accuracy rate of 73% remains high for nationwide analyses across all areas. For comparison, the Corine Land Cover project, which has an international scope, reports an accuracy of 67.8% in certain countries such as Norway, with higher error rates in rural areas [41]. In some scenes where the radiometric characteristics of the landscape are fully represented in the training dataset and the image quality is high (Figure 8), the accuracy of urban footprint detection can reach up to 91%.

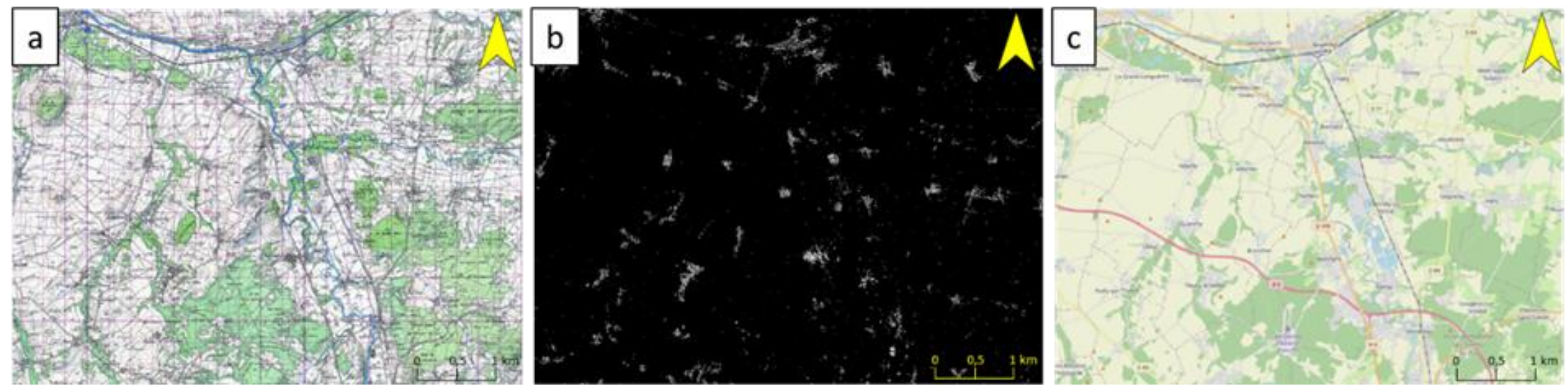

**Figure 8**. (a) Original scan Histo map , (b) predicted urban footprints, (c) Openstreet map "North of Auxerre city"

In more rural areas—particularly agricultural zones with linear textures and internal dark patterns—the model encounters greater difficulty in eliminating false positives (Figure 9) and even retains some residual text elements (Figure 9b). This results in a decrease in overall accuracy, dropping to levels between 45% and 60%.

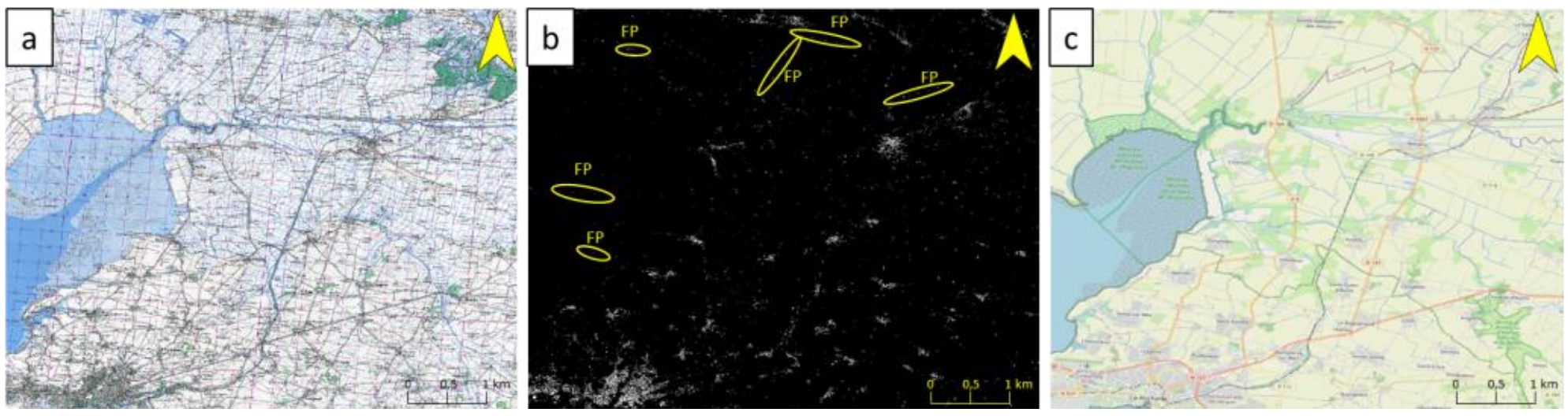

**Figure 9**. (a) Original scan Histo map , (b) predicted urban footprints, (c) Openstreet map "North east of La Rochelle city".

Since the predicted urban footprint scenes have a resolution of 5 m (matching that of the native historical scan maps), they can also undergo resampling to resolutions more commonly used in international datasets, such as GHSL at 100 m [42], or Landsat MSS with scenes from the early 1970s at 80 m resolution [43]. This resampling process can help remove many isolated residual false positives, particularly in rural areas, through the application of a majority-based elimination algorithm.

At the scale of the French territory, the overall accuracy rate (73%) can be explained by the existence of two distinct zones (Figure 10). An East–Central zone (with some tiles in the West) exhibits very high quality (up to 85%). In contrast, a Western zone—characterized by scenes with highly heterogeneous tones and numerous residuals from text and roads—shows a relatively low rate, ranging from 40% to 60%.

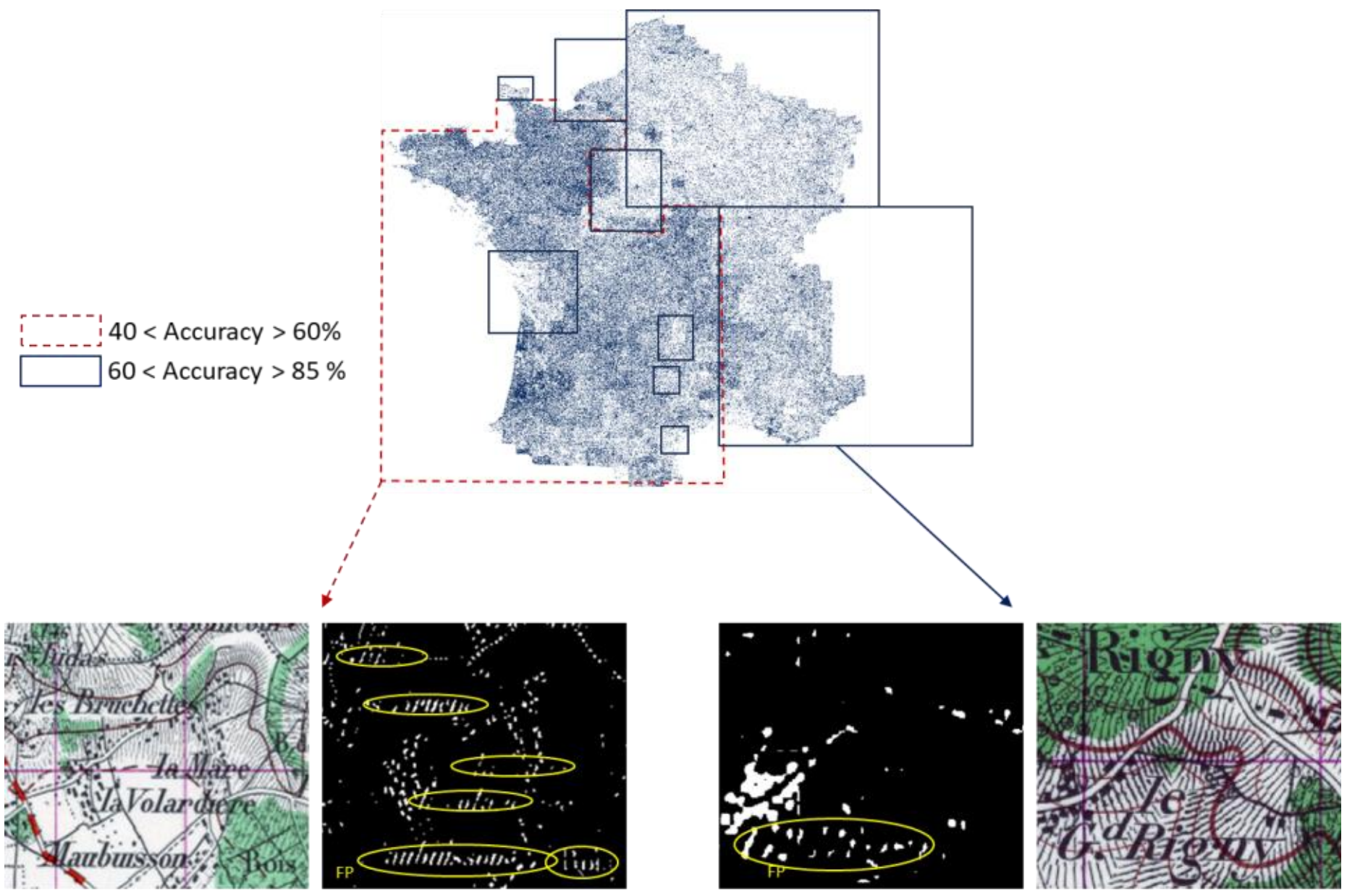

**Figure 10**. Global quality (national scale) of the predicted urban footprint

At the city scale level (Figure 11), the prediction is highly accurate, particularly in peripheral areas (since the datasets were produced from predominantly urban samples). Thus, the prediction is truly based on the intelligence developed by the model.

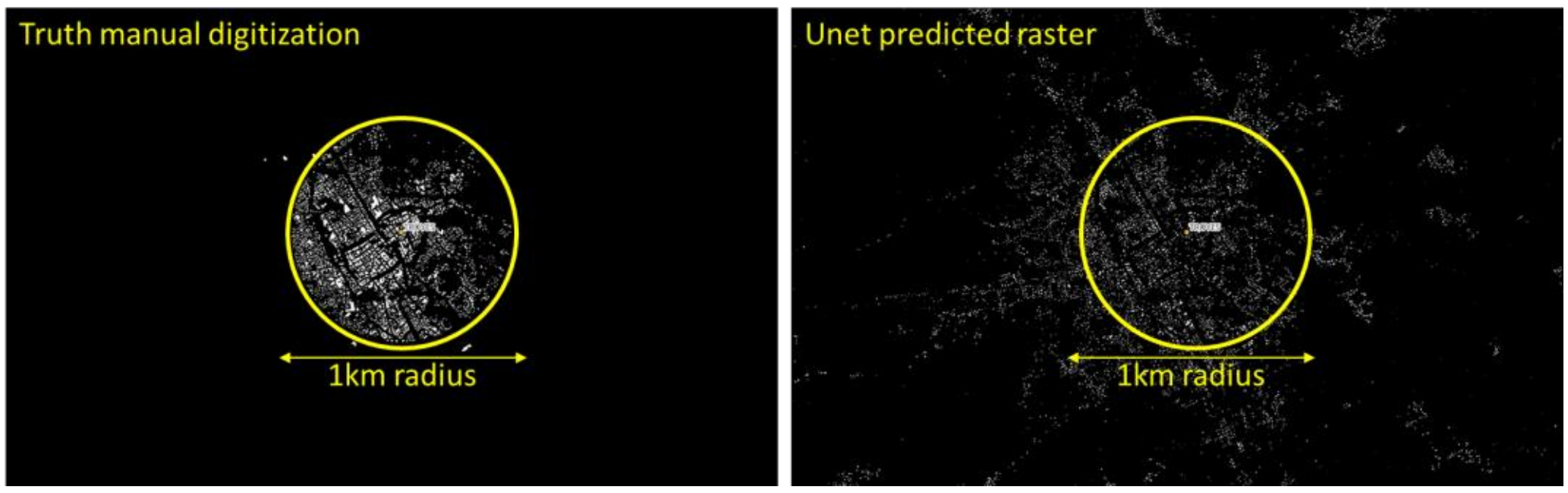

**Figure 11.** Truth urban footprints for Troyes city (center area) vs predicted raster (city and suburbs)

It is nevertheless noted that urban density is slightly lower in the predicted scenes compared to the original dataset (due to intra-urban polygon heterogeneity, the effect of pixelization, and the preservation of maps).

*4.1 Thematic application*

The produced urban footprint layer, made freely available, offers a wide range of potential applications in geography. Resampled to a 100 m resolution to ensure compatibility with the GHSL database [42], this will allow at least to visualize multitemporal urban sprawl. Preliminary visual assessments for several French cities already indicate substantial potential for analysis (Figure 12), particularly in highlighting the extensive urban sprawl of Paris and the markedly high sprawl observed in Marseille.

.

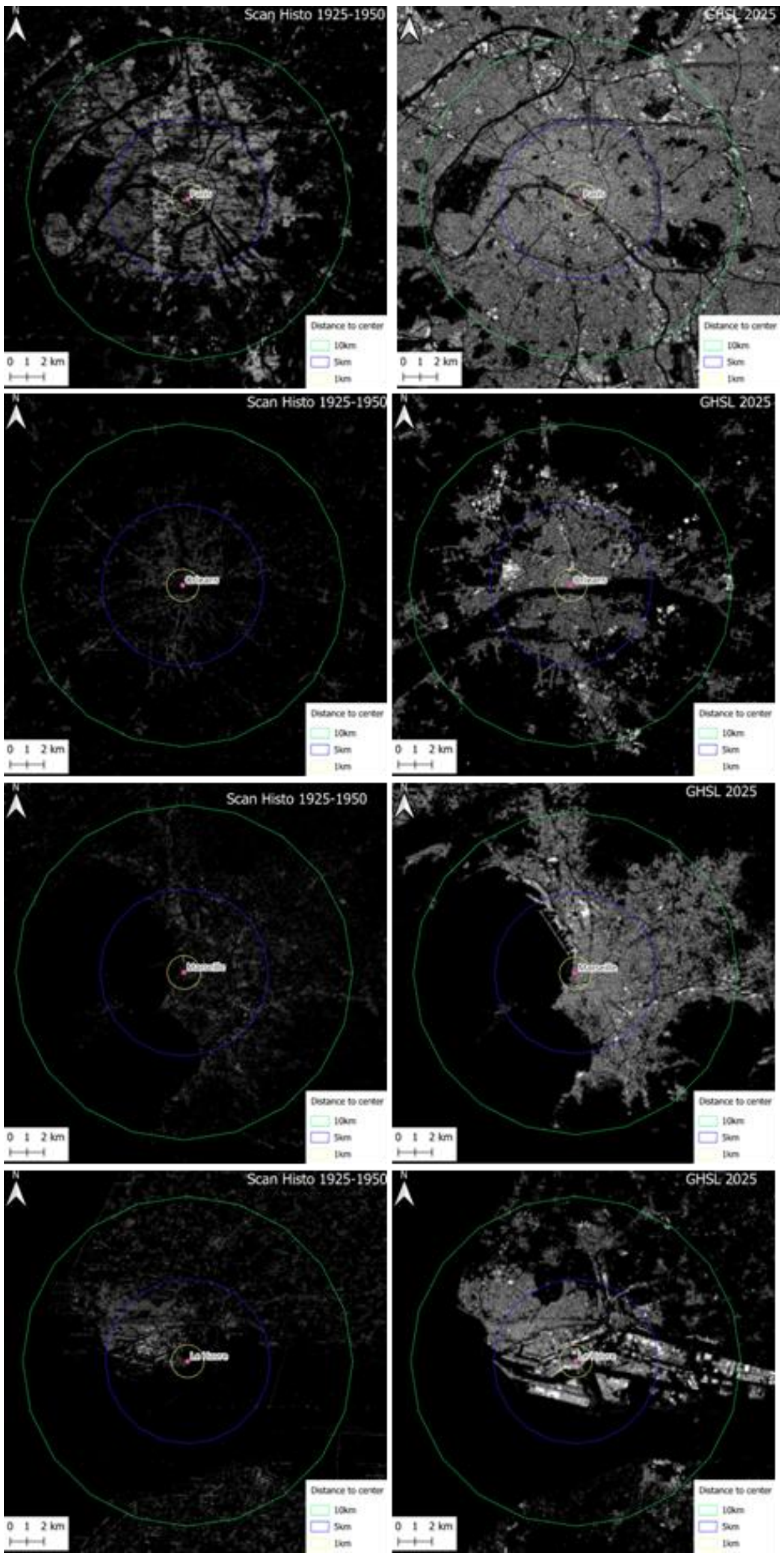

**Figure 12**. Overview of urban sprawl (Scan Histo vs GHSL)

A significant pattern of urban growth is also observed for the city of Orléans, particularly along the Loire River, while in the case of Le Havre, the expansion of its port appears to accompany the broader process of urbanization. These preliminary observations highlight the potential of this layer for scientific use, for example to compare the extent of urban areas before and after the Second World War. It is worth noting that the majority of publications on quantitative measurements of urban sprawl date back to the early 1970s, coinciding with "the advent of the first Landsat MSS imagery over France.

## 5. Conclusions

This work demonstrates a practical pipeline for semantic segmentation of urban areas from satellite imagery using a customized deep learning approach. The trained UNet model, combined with a geospatially-aware inference system, enables fast and automated urban mask generation at scale. Future improvements will include multi-class segmentation, performance benchmarking, and optical character recognition (OCR).

This study reveals the feasibility of extracting coherent historical urban footprints from archival maps using a deep learning approach. The method leverages the richness of the Scan Histo archive and provides a scalable framework applicable to other historical datasets. Future work will explore temporal generalization and integration with earlier and later map series to build continuous historical urban datasets (Etat major maps of the 19th century and Cassini maps for the 18th).

Although some minor anomalies persist, the approach produces a promising pre-trained model capable of overcoming color tone variations, the majority of textual elements and roads, and the heterogeneity of urban objects.

The cascade "encoder-decoder" architecture of UNet, when trained on a relevant dataset that captures the diverse patterns urban objects can take, also makes it possible to eliminate non-representative objects sharing similar textures (such as pictograms of airports, ports, castles etc.). This is particularly encouraging, as it reflects a level of understanding often associated with human cognition.

The results are openly shared,and a database containing ground-truth shapefiles is already available online, along with the code, the resulting raster, and the trained model. This open-access approach is intended to enable other users to contribute improvements.

**Author Contributions:** Conceptualization, W.R.; formal analysis, W.R.; methodology, W.R.; writing—original draft preparation, W.R.; funding acquisition, R.L., M.L.T.; supervision, R.L.; project administration, R.L., M.L.T.; writing—review and editing, R.L., M.L.T. All authors have read and agreed to the published version of the manuscript.

**Funding:** This research was funded by Normandy Region (https://www.normandie.fr/aides-regionales), and "The APC was funded by PLACES Lab (CY Cergy Paris University)".

**Data Availability Statement:** All the data (base maps and results) , we also provide the codes and learning dataset and the trained model in openacess:

Basemaps: https://geoservices.ign.fr/telechargement-api/SCAN50-HISTORIQUE

Results: https://nakala.fr/10.34847/nkl.aea7388x

Codes: https://github.com/wilius47/UNET_for_Historical_map/tree/main/Code

Pretrained Model: https://nakala.fr/10.34847/nkl.aea7388x

Datasets ready: https://api.nakala.fr/data/10.34847/nkl.aea7388x/09f4d3991421c34d633fbe620045dddb937adc27

Datasets in shapefile: https://nakala.fr/10.34847/nkl.aea7388x

Technical paper on datasets production: https://nakala.fr/10.34847/nkl.aea7388x#59f9dd5722144173b394671ad30087ba8fd3515d

**Acknowledgments:** The authors would like to thank Julien Perret from IGNF, and Alexis Lietvin from the University of Cambridge, for the scientific discussions on historical map knowledge and data sources. The authors have reviewed and edited the output and take full responsibility for the content of this publication.". The authors thanks also the Normandy Region for funding the SUCHIES project (RIN 2021), and the CRIANN "Regional Center for Computing and Digital Applications of Normandy", for allowing the use of AUSTRAL Server.

**Conflicts of Interest:** The authors declare no conflicts of interest. The funders had no role in the design of the study; in the collection, analyses, or interpretation of data; in the writing of the manuscript; or in the decision to publish the results

## Abbreviations

SCAN HISTO : Historical Scanned map of the 20th century (1925-1950)

CRIANN: Normandy Regional Center for Computing and Digital Applications

Austral-HPDA : server name –High-Performance Data Analytics

SUCHIES: Structured urban database over time and scales (authors research project)

IGN/IGNF: French Geographic & forest institute (national map producer)

GHSL : Global Human Settlement Layer (open access urban footptints database)

**About authors**

**Walid Rabehi** – Associate Professor in Geography at CY Cergy Paris University; affiliated with PLACES research lab. Specializes in remote sensing, spatial analysis, machine learning–based land-use modeling, and coastal vulnerability; recent work includes deep-learning approaches to coastal erosion and SDG 15.3.1 land-cover monitoring. He also was a research engineer of SUCHIES project.

**Rémi Lemoy** – Associate Professor of Geography at the University of Rouen (affiliated to UMR IDEES Lab). Background in physics, economics, and geography; expert in urban modeling using statistical physics, urban economics, and agent-based simulations (co-leads SUCHIES project). He is also winner of the 2020 Michael Breheny Prize for work on scaling of urban forms.

**Marion Le Texier** – Associate Professor in Geography at Université Paul Valéry – Montpellier 3 and (affiliated to LAGAM lab). Her research focuses on systemic and multi-scale analysis of territorial dynamics, including mobility, transport infrastructure, location of activities and households, and land use. She also works on environmental equity and risk communication, co-leads the SUCHIES project, and contributes to editorial initiatives and thematic leadership in complex systems modeling.

**Supplementary Materials:**;

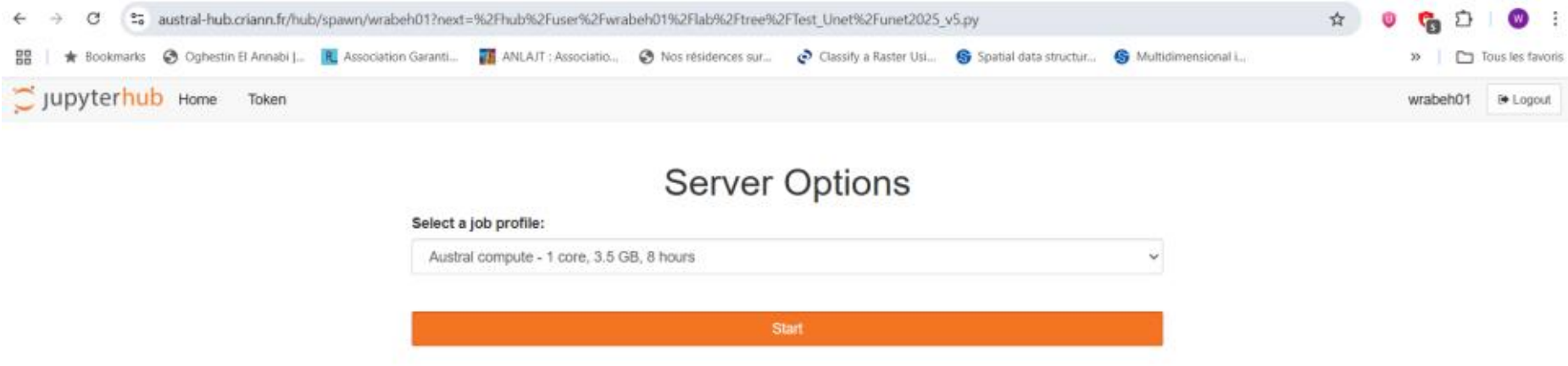

Figure S1: Austral sever access

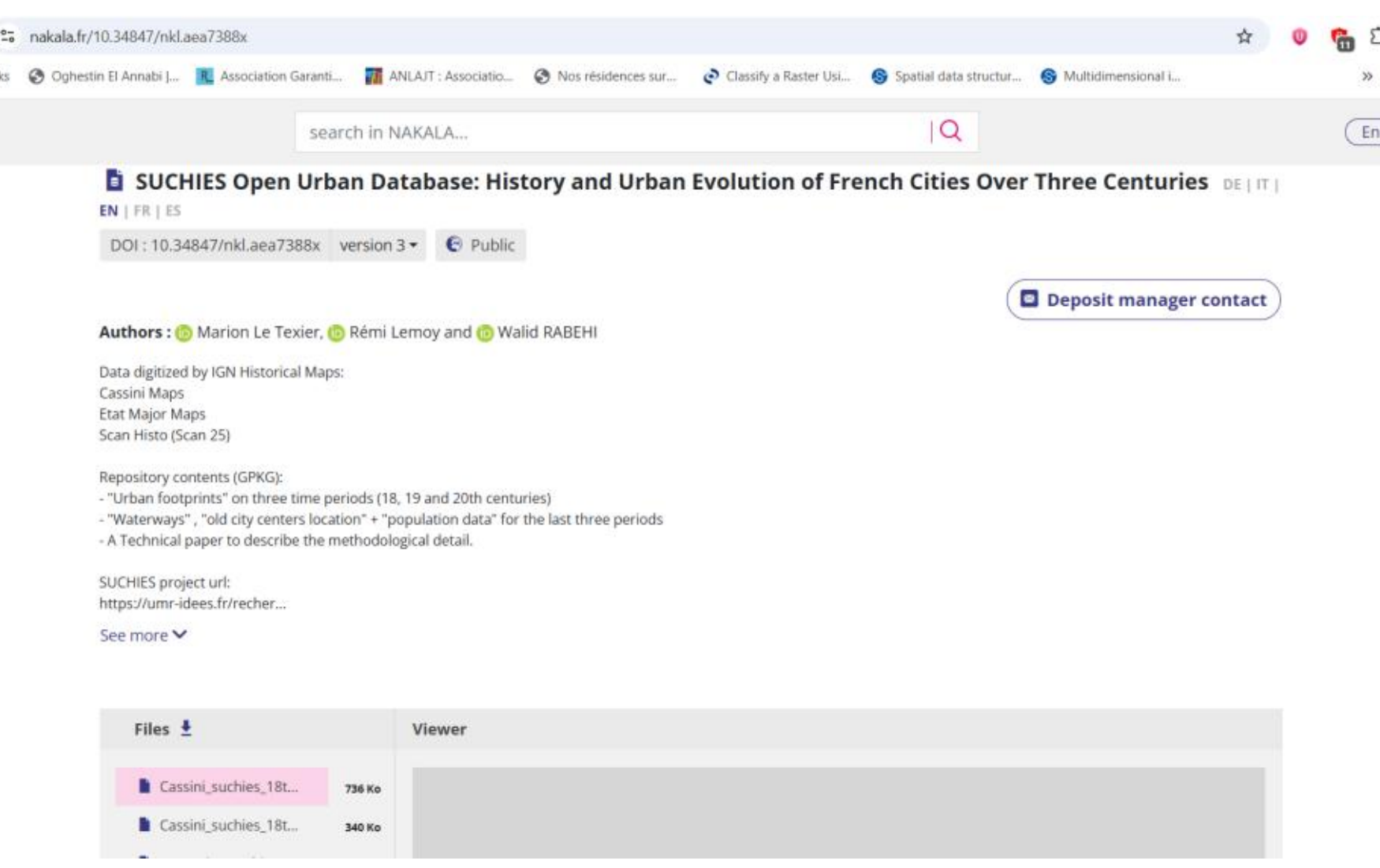

Figure S2: SUCHIES openaccess database

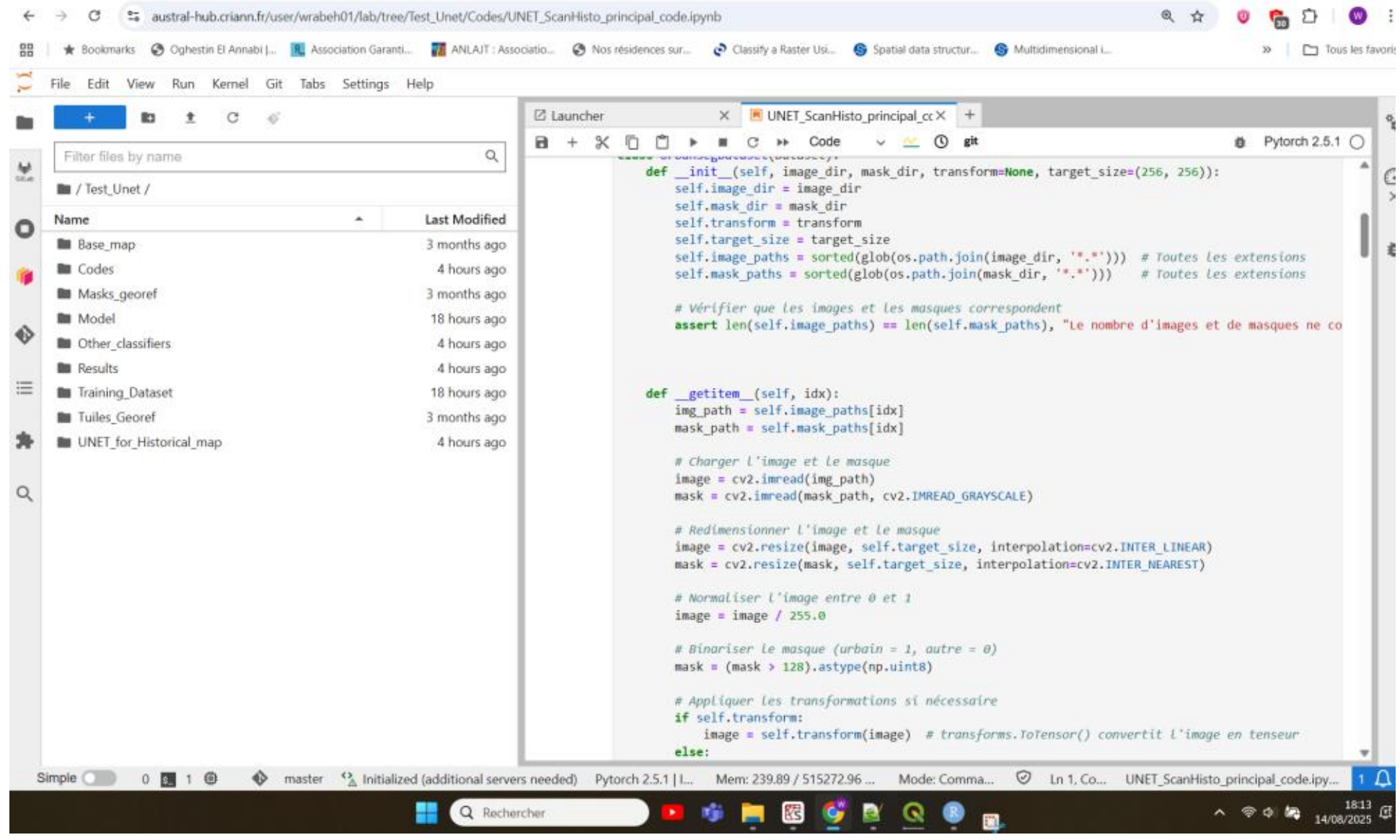

Figure S3: Austral notebook coding

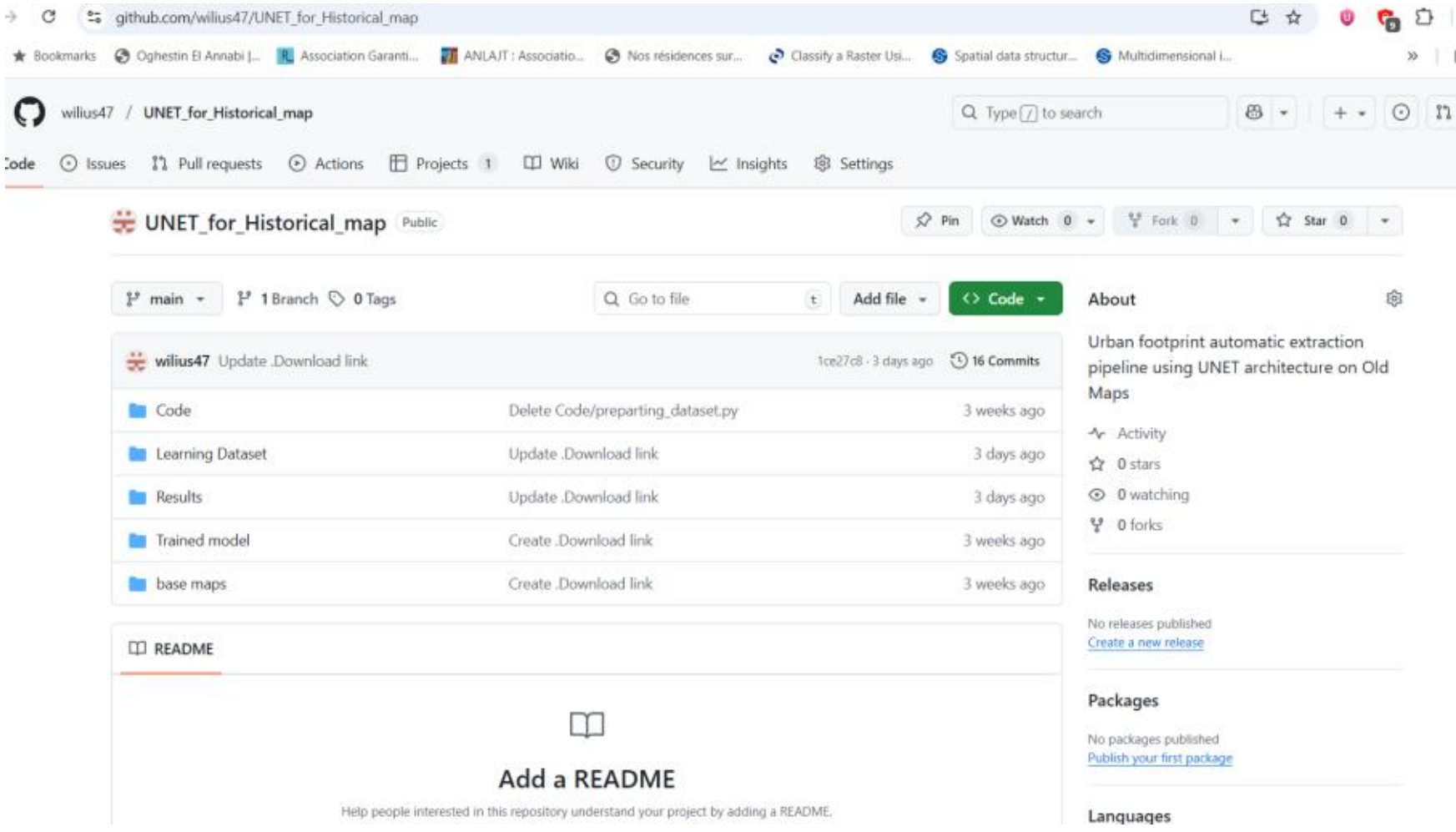

Figure S4: Github account of the project.